\documentclass[letterpaper]{article} 

\usepackage{amsmath,amsfonts,bm}

\def\eqref#1{equation~\ref{#1}}

\def\1{\bm{1}}

\def\vx{{\bm{x}}}
\def\vy{{\bm{y}}}
\def\vz{{\bm{z}}}

\def\mE{{\bm{E}}}

\def\mH{{\bm{H}}}

\def\mS{{\bm{S}}}

\DeclareMathAlphabet{\mathsfit}{\encodingdefault}{\sfdefault}{m}{sl}
\SetMathAlphabet{\mathsfit}{bold}{\encodingdefault}{\sfdefault}{bx}{n}

\usepackage[]{aaai24}  
\usepackage{times}  
\usepackage{helvet}  
\usepackage{courier}  
\usepackage[hyphens]{url}  
\usepackage{graphicx} 
\usepackage{natbib}  
\usepackage{caption} 
\usepackage{algorithm}
\usepackage{algorithmic}

\usepackage{booktabs}
\usepackage{multirow}

\usepackage{newfloat}
\usepackage{listings}
\DeclareCaptionStyle{ruled}{labelfont=normalfont,labelsep=colon,strut=off} 
\floatstyle{ruled}
\newfloat{listing}{tb}{lst}{}
\floatname{listing}{Listing}
\title{An Autoregressive Text-to-Graph Framework for Joint \\ Entity and Relation Extraction}

\author{
    Urchade Zaratiana\textsuperscript{1,2},
    Nadi Tomeh\textsuperscript{2},
    Pierre Holat\textsuperscript{1,2},
    Thierry Charnois\textsuperscript{2}
}

\affiliations{
    \textsuperscript{1} FI Group, Puteaux, France \\ \textsuperscript{2} LIPN - Université Sorbonne Paris Nord - CNRS UMR 7030, Villetaneuse, France\\
    \{zaratiana, tomeh, charnois\}@lipn.fr, pierre.holat@fi-group.com
}

\usepackage{bibentry}

\begin{document}

\maketitle

\begin{abstract}
In this paper, we propose a novel method for joint entity and relation extraction from unstructured text by framing it as a conditional sequence generation problem. In contrast to conventional generative information extraction models that are left-to-right token-level generators, our approach is \textit{span-based}. It generates a linearized graph where nodes represent text spans and edges represent relation triplets. Our method employs a transformer encoder-decoder architecture with pointing mechanism on a dynamic vocabulary of spans and relation types. Our model can capture the structural characteristics and boundaries of entities and relations through span representations while simultaneously grounding the generated output in the original text thanks to the pointing mechanism. Evaluation on benchmark datasets validates the effectiveness of our approach, demonstrating competitive results. Code is available at https://github.com/urchade/ATG.
\end{abstract}

\section{Introduction}
Joint entity and relation extraction is a fundamental task in Natural Language Processing (NLP), serving as the basis for various high-level applications such as Knowledge Graph construction \citep{ye-etal-2022-generative} and question answering \citep{chen-etal-2017-reading}.
Traditionally, this task was tackled via pipeline models that independently trained and implemented entity recognition and relation extraction, often leading to error propagation \citep{10.1007/10704656_11}. Deep learning has led to the creation of end-to-end models, allowing for the use of shared representations and joint optimization of loss functions for both tasks \citep{wadden-etal-2019-entity, wang-lu-2020-two, Zhao_Yan_Cao_Li_2021,zhong-chen-2021-frustratingly, yan-etal-2021-partition}. Despite this advancement, these models essentially remain pipeline-based, with entity and relation predictions executed by separate classification heads, thereby ignoring potential interactions between these tasks.

Recent advancements have seen a shift towards ``real'' end-to-end solutions, where the prediction of entities and relations is intertwined, accomplished through autoregressive models. These models treat the joint entity-relation task as a process of generating plain text, employing \textit{augmented languages} to encode and decode structural information \citep{paolini2021structured, lu-etal-2022-unified, liu-etal-2022-autoregressive, fei2022lasuie}. While these models have achieved remarkable performance, we argue that they also expose room for improvement, especially in terms of grounding the output in the input text.

In this paper, we present an autoregressive transformer encoder-decoder model that generates a linearized graph instead of generating plain text. Our model makes use of a pointing mechanism \citep{vinyals2017pointer} on a dynamic vocabulary of spans and relations, providing explicit grounding in the original text. In fact, without grounding, models can generate output that are semantically coherent but contextually detached from the input. Our pointing mechanism mitigates this issue by ensuring that the decoder's outputs, specifically the entity spans, are directly tied to the input text. Furthermore, by generating spans and relations directly from the text, rather than producing standalone plain text, our model encode the structural characteristics and boundaries of entities/spans more accurately, which can be missed by previous generative information extraction models. The cornerstone of our solution is the explicit enumeration of all spans\footnote{Up to a certain length in practice.} at the encoder's output, making them readily available to the decoder. Although the number of spans can be extensive, we note that when bounding the span size, our model's vocabulary is typically smaller than that of traditional language models \citep{Devlin2019BERTPO,Raffel2019ExploringTL}, as discussed in subsequent sections.

\begin{figure*}[h]
\centering
\includegraphics[width=1.\textwidth]{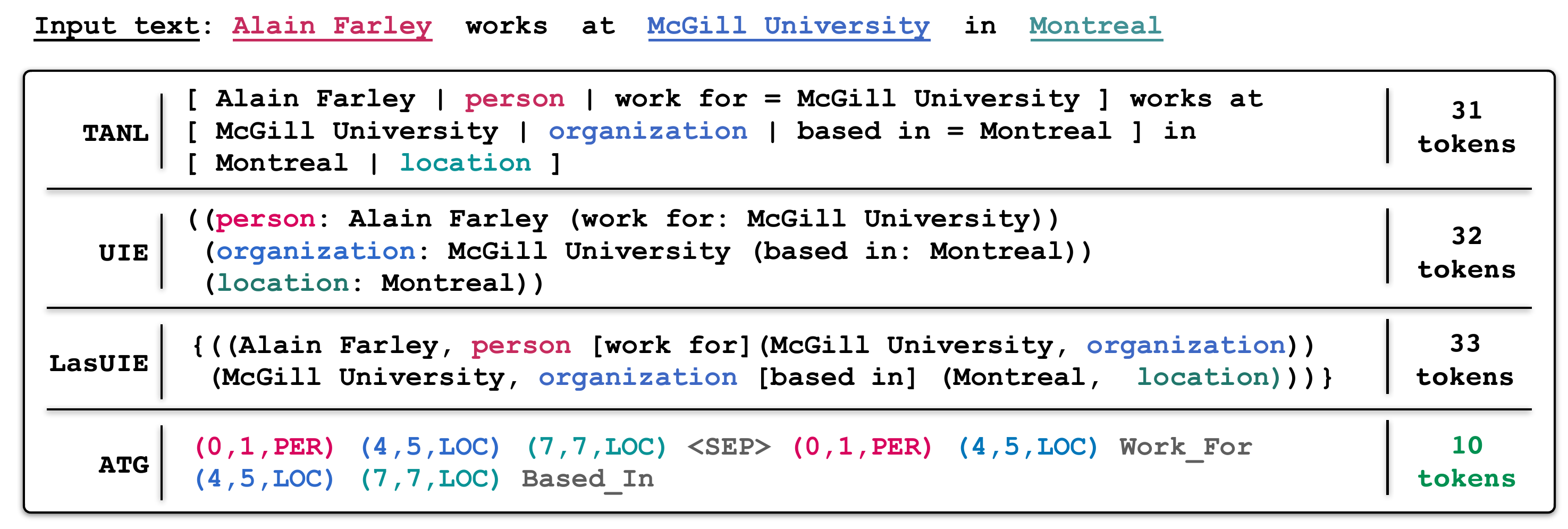}
\caption{Linearization for different models. In contrast to existing approaches (\textit{TANL} \citep{paolini2021structured}, \textit{UIE} \citep{lu-etal-2022-unified}, \textit{LasUIE} \citep{fei2022lasuie}), our proposed model, \textit{ATG}, generates spans (along with relation/special tokens) instead of text tokens, which allows for a shorter output sequence, richer (span-level) representation and fully controlled decoding.}
\label{lincom}
\end{figure*}

Moreover, as previous generative IE models operate at the token level, they scatter the information regarding an entity's span and its boundaries over multiple decoding steps. In contrast, generating an entity and its type in our approach is accomplished in a single decoding step, resulting in shorter sequence (Figure \ref{lincom}). Additionally, our method naturally ensures the well-formedness of the output while some generative IE models that produce text, often require non-regular constraints. As an example, in TANL \cite{paolini2021structured}, if a portion of the generated sentence has an invalid format, that segment is discarded. Such issues are readily addressed in our model since the vocabulary can be fully controlled.

\begin{figure*}
    \centering
    \includegraphics[width=\textwidth]{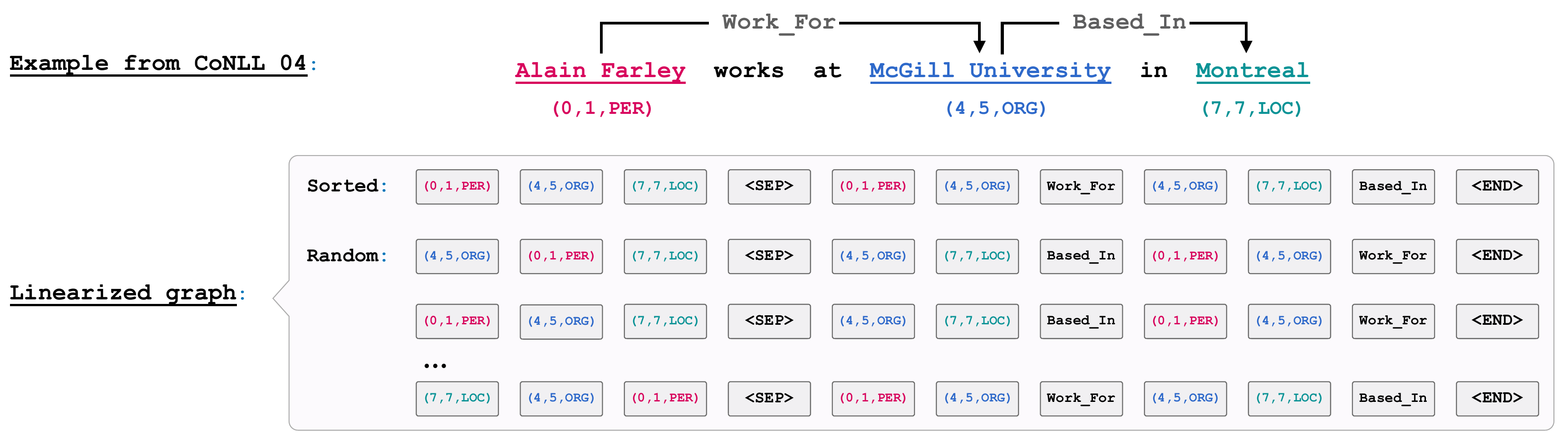}
    \caption{Linearization for Information Graph Generation. The input text is mapped into an information extraction graph. The graph consists of entities and relation triplets, which are generated sequentially by first producing entity spans (represented by start word, end word, and entity type) followed by relation triplets (head entity, tail entity, and relation type).}
    \label{fig:graph_gen}
\end{figure*}

We evaluated our model on three benchmark datasets for joint entity and relation extraction: CoNLL 2004, SciERC, and ACE 05. Our model demonstrated competitive performance on all datasets. Our contributions can be summarized as follows:

\begin{itemize}
    \item We propose a novel method for joint entity and relation extraction by framing it as a conditional sequence generation problem. Our approach generates a linearized graph representation, where nodes represent entity spans and edges represent relation triplets.
    \item Our model employs a transformer encoder-decoder architecture with a pointing mechanism on a dynamic vocabulary of spans and relation types. This allows to capture the structural characteristics and boundaries of entities and relations while grounding the generated output in the original text.
    \item We demonstrate the effectiveness of our approach through extensive evaluations on benchmark datasets, including CoNLL 2004, SciERC, and ACE 05. Our model achieves state-of-the-art results on CoNLL 2004 and SciERC, surpassing previous comparable models in terms of Entity F1 scores and Relation F1 scores.
\end{itemize}

\section{Task Definition}

We address the task of joint entity and relation extraction from text as a graph generation approach. Our proposed model generates nodes and edges as a single sequence, effectively integrating both entity and relation extraction into a unified framework. Formally, the task can be defined as follows: Given an input text sequence $\vx = \{x_1, x_2, ..., x_L\}$, where $x_i$ represents the $i$-th token in the sequence, our objective is to generate a linearized graph representation $\mathbf{y} = \{y_1, y_2, ..., y_M\}$, where $y_j$ represents a token in the generated sequence. As shown in Figure \ref{fig:graph_gen}, in Each token $y_j$ can take one of three forms:
\begin{itemize}
\item \textbf{Entity span}: The token $y_j$ represents an entity span, defined as $y_j = (s_j, e_j, t_j)$, where $s_j$ and $e_j$ denote the starting and ending positions of the entity span, and $t_j$ denotes the type of the entity.
\item \textbf{Relation type}: The token $y_j$ represents a relation type between two entities, such as \texttt{Work\_For} relation.
\item \textbf{Special token}: The token $y_j$ represents special tokens used in the generation process, such as \texttt{<SEP>} to separate entities and relations or the \texttt{<END>} to stop the generation.
\end{itemize}

\begin{figure*}[]
    \centering
    \includegraphics[width=\textwidth]{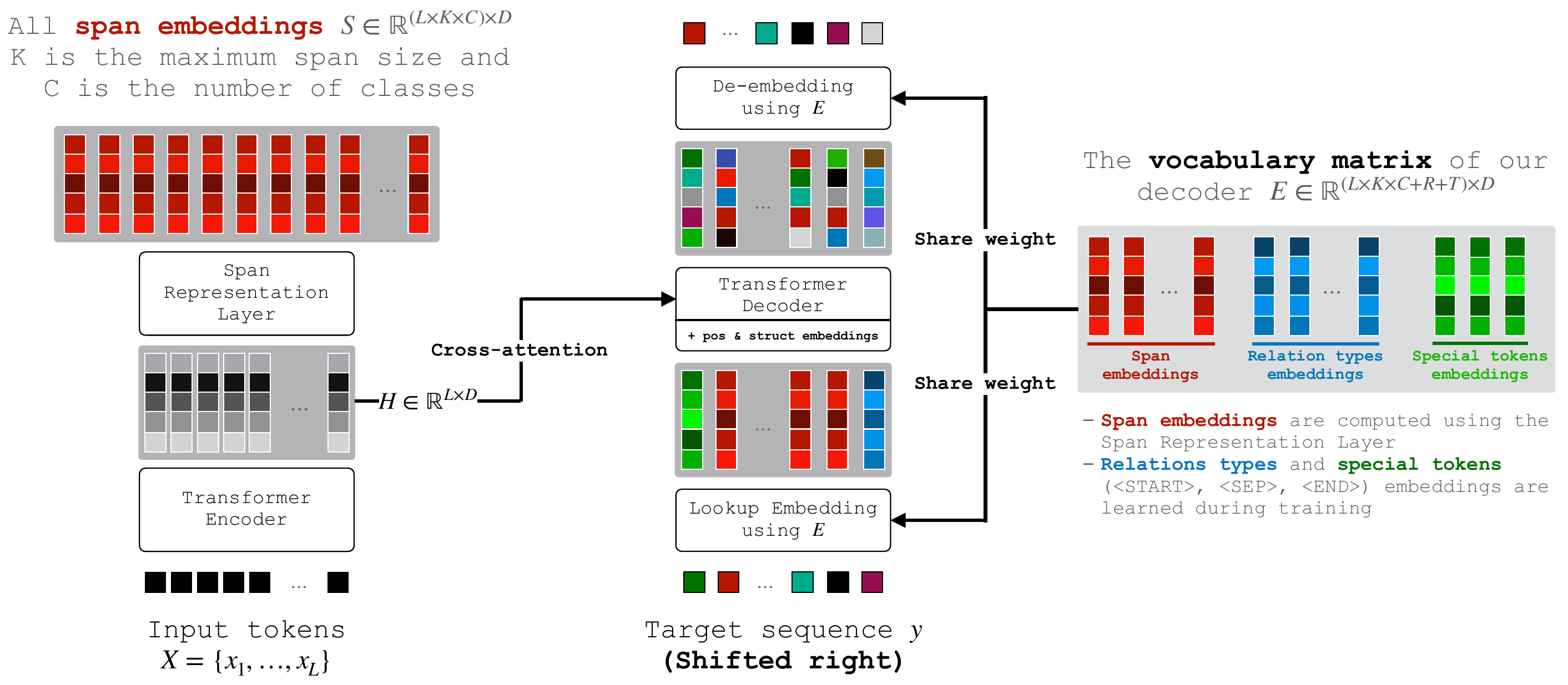}
    \caption{Illustration of the architecture of our model, ATG. (\textit{left}) The Encoder takes in the input sequence $X$ and generates representations of the tokens $\mH$ and spans $\mS$. (\textit{middle}) The Decoder then generates the next token conditioned on the previous tokens and the input representation $\mH$. (\textit{right}) The vocabulary matrix used for decoding consists of the concatenation of span embeddings $\mS$, learned relation type embeddings, and special token embeddings.}
    \label{fig:model_arch}
\end{figure*}

The template employed in our model, we refer to \textbf{ATG} (\textbf{A}utoregrestive \textbf{T}ext-to-\textbf{G}raph), is depicted in Figure \ref{fig:graph_gen}. It starts by generating the entities, followed by a \texttt{<SEP>} token, and ends with the relation triplets, each consisting of head node, tail node and edge/relation type. During training, we try two distinct orderings for the graph linearization: \textit{sorted} and \textit{random}. As shown in the Figure \ref{fig:graph_gen}, the \textit{sorted} linearization organizes entities and relations based on their positions in the original text, while the \textit{random} linearization randomly shuffles entities and relations order.

\section{\label{sec:archi}Model Architecture}

Our model, ATG, employed an encoder-decoder architecture, which processes the input text sequence and produces a linearized graph as illustrated in the Figure \ref{fig:model_arch}. 

\subsection{Encoder}

The encoder in ATG utilizes a transformer layer that takes an input text sequence $\mathbf{x}$ and outputs token representations $\mathbf{H} \in \mathbb{R}^{L \times D}$, where $D$ is the model dimension.

\subsection{Vocabulary Construction}

\paragraph{Dynamic vocabulary} To enable the pointing mechanism in our decoder, we construct a dynamic vocabulary matrix $\mathbf{E}$ that includes embeddings for spans, special tokens, and relation types. While special tokens and relation type embeddings are randomly initialized and updated during training, the span embeddings are dynamically computed \citep{zaratiana-etal-2023-filtered}, \textit{i.e} their representations depend on the input sequence. More specifically, the embedding of a span $(\text{start}, \text{end}, \text{type})$ is computed as follows:
\begin{align}
\mathbf{S}_{[\text{start}, \text{end}, \text{type}]} = \mathbf{W}_{\text{type}} ^T [\mathbf{h}_{\text{start}} \odot \mathbf{h}_{\text{end}}]
\end{align}
In this equation, $[\odot]$ represents a concatenation operation; $\mathbf{h}_{\text{start}}$ and $\mathbf{h}_{\text{end}}$ denote the representations of tokens at the start and end positions, respectively. $\mathbf{W}_{\text{type}} \in \mathbb{R}^{2D\times D}$ is a weight matrix associated with the entity type (\textit{i.e}, there is a $\mathbf{W}_{\text{type}}$ for each entity types in a datasets). Finally, the vocabulary embedding matrix $\mathbf{E}$ is formed by stacking all the span embeddings $\mathbf{S}$, special token embeddings $\mathbf{T}$, and relation type embeddings $\mathbf{R}$.

\paragraph{Vocabulary size \label{sec:vocsize}} The size of the vocabulary matrix $\mathbf{E} \in \mathbb{R}^{V \times D}$ is $V=L \times K \times C + R + T$, where $L$ represents the sequence length, $K$ the maximum span size, $C$ the number of entity types, $R$ the number of relations, and $T$ the number of special tokens (\texttt{<START>}, \texttt{<END>}, and \texttt{<SEP>}).
Let's take the CoNLL 2004 dataset as an example to illustrate this. This dataset has the following characteristics: $K=12$, $C=4$, $R=5$, and $T=3$. Considering a sentence of length 114 (which is the maximum length in the training set), the resulting vocabulary size would be 5480. This size is considerably smaller when compared with the vocabulary size of a typical language model, which usually hovers around 30,000 distinct tokens.

\begin{figure*}[t]
    \centering
\includegraphics[width=\textwidth]{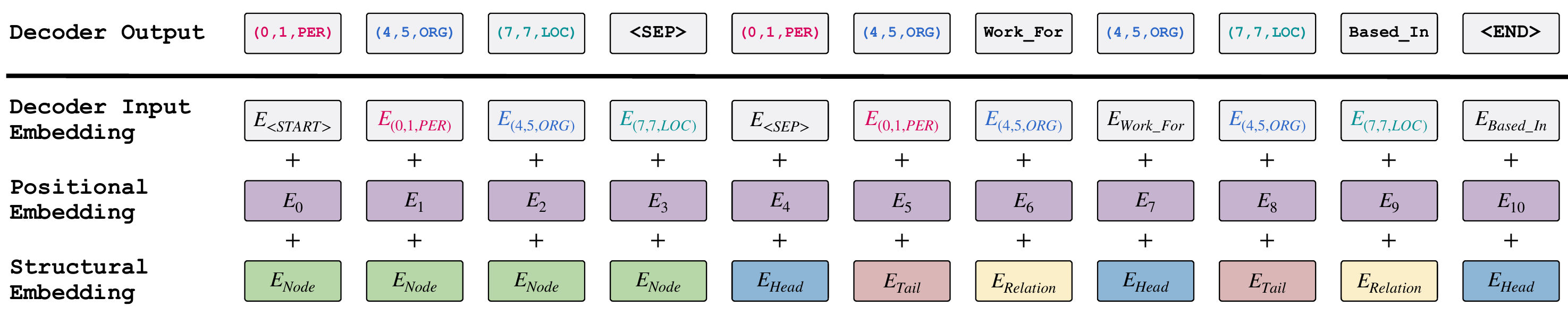}
    \caption{Input/ouptut of the decoder. The process starts with the special token \texttt{<START>} and continues until the \texttt{<END>} token is generated. To separate the generation of nodes and edges, a special token \texttt{<SEP>} is used. At each position, the decoder takes in the sum of the embedding of the current token, absolute position embedding, and structural embedding.}
    \label{fig:embed}
\end{figure*}

\subsection{\label{sec:decoder}Decoder}
The decoder is a causal transformer trained to predict the next token in the sequence, akin to traditional language modeling. However, it is important to note that the vocabulary of our decoder consists of entity spans, relation types, and special tokens, rather than plain text. The decoder conditions its predictions on the previously generated tokens $y_{<j}$ using self-attention and on the input token representations $\mathbf{H}$ using cross-attention. This enables the decoder to attend to relevant information from both the previously generated tokens and the input text.  Through attention visualizations, as depicted in Figure \ref{fig:selfatt} and \ref{fig:crossfig}, we observed that the model effectively harnesses both sources of information. Finally, The training objective aims to maximize the following conditional probability:
\begin{equation}
  p(\vy|\vx) = \prod_{j=1}^{M} p(y_j | \vy_{<j}, \mH)  
\end{equation}

This is achieved during the training by minimizing the negative log-likelihood of a reference sequence obtained by linearizing the reference IE graph.
Details about the decoder input and output are given in the subsequent paragraphs.

\paragraph{Decoder input embedding} The embedding step feeds the previous decoder outputs $y_1, \ldots, y_{i-1}$ into the model using the vocabulary matrix $\mathbf{E}$, along with \textit{positional} and \textit{structural} embeddings as shown in Figure \ref{fig:embed}. This process can be expressed as follows:
\begin{equation}
    \begin{split}
    \vz_1, \ldots, \vz_{i-1} =& \mE[y_1, \ldots, y_{i-1}] \\+& \mE_{pos}[1, \ldots, i-1]
    \\+& \mE_{struct}[y_1, \ldots, y_{i-1}]
\end{split}
\end{equation}

Here, $\mE[y_1, \ldots, y_{i-1}]$ corresponds to the token embeddings, which may corresponds to spans, relation types, or special tokens embeddings depending on the nature of $y_{<i}$. The matrix $\mE_{pos}$ represents absolute positional embedding. It allows to capture the positional information of the decoder outputs from $y_1$ to $y_{i-1}$. Additionally, the structural embedding $\mE_{struct}$ serve as indicators to guide the model in generating specific elements. In particular, they provide information about whether the model should generate a Node (before \texttt{<SEP>} tokens), Tail, Head nodes, or Relation types (as illustrated in Figure \ref{fig:embed}). Both the absolute position encoding and the structural embedding are randomly initialized and updated during training. In summary, by combining these embeddings, ATG can capture the semantic, positional and structural information about the linearized graph.

\begin{figure}[!b]
    \centering
    \includegraphics[width=1.\columnwidth]{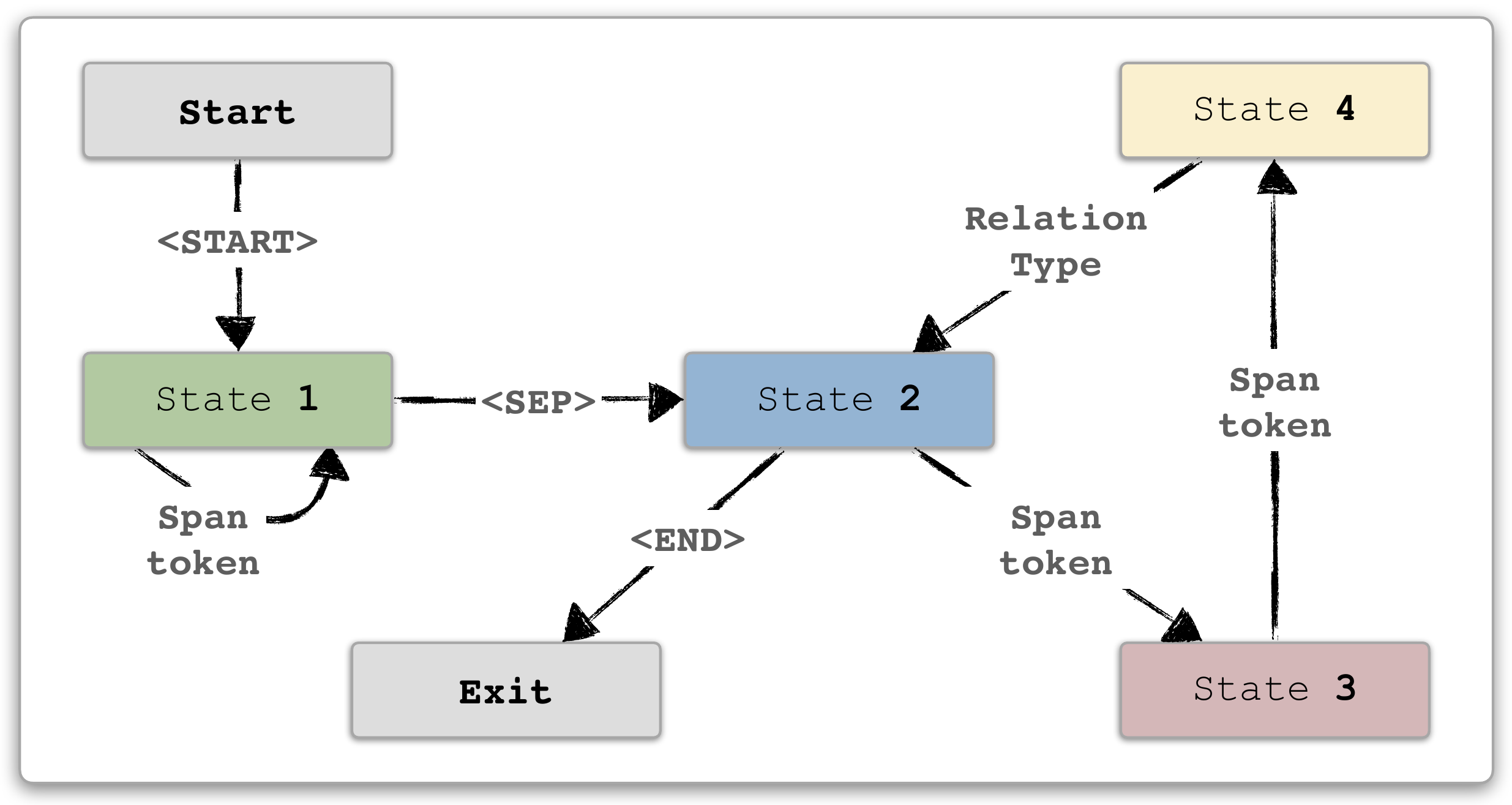}
    \caption{State-Transition diagram for constrained decoding. This diagram illustrates the state-based decision process used during the inference phase, which ensures the generation of a correct graph. Each state is represented by a node, and directed edges indicate valid actions. We use the same color code as the \textit{structural embedding} in Figure \ref{fig:embed}.}
    \label{fig:state_approach}
\end{figure}

\paragraph{Decoder output} We define $\tilde{\vz_i}$ as the hidden state at the last position of the decoder output sequence obtained by feeding the previous output embedding and the encoder outputs $\mH$ (for cross-attention) to the decoder, \textit{i.e},
\begin{align}
    \tilde{\vz_i} = \mathcal{D}ecoder(\vz_1, \ldots, \vz_{i-1}; \mH)\mathbf{[-1]}
\end{align}

Then, to compute the probability distribution over the dynamic vocabulary for generating the next token, $y_i$, our model employs the softmax function on the dot product between the dynamic vocabulary embedding matrix $\mE$ and $\tilde{\vz_i}$:
\begin{align}
    p(y_i | \vy_{<i}, \mH) = \frac{\exp \mE^T\tilde{\vz_i}}{\sum_{k=1}^V (\exp \mE^T\tilde{\vz_i})_k}
\end{align}

The probabilities generated by this formulation allow the model to select the appropriate token from the vocabulary for generating a span, special token, or relation type.

\paragraph{Constrained decoding} During inference, we sample from the model by enforcing constraints that preserve the well-formedness of the output graph. More specifically, during inference, we feed our model with the \texttt{<START>} token, and the generation process is guided by state-transition constraints as outlined in Figure \ref{fig:state_approach}. This structured approach ensures that each step in the generation aligns with the defined template, thereby maintaining the well-formedness of the output and allowing the production of valid IE graphs. All generations start with the start token \textbf{Start} state and continue until the \texttt{<END>} token is sampled. In practice, we also add other constraints: in state \textbf{1}, we prevent the repetition of already generated spans, and in state \textbf{3}, we ensure the tail span is different from the head. Furthermore, it is also possible to incorporate domain knowledge into the prediction. For instance, if the type of a head entity is PER and the tail entity is ORG, the relation can be constrained to be Work\_For (CoNLL 04 dataset).

\subsection{Training with Sentence Augmentation\label{sent:aug}}
In our work, we observed oversmoothing \citep{kulikov-etal-2022-characterizing}, where the model prematurely generates the \texttt{<EOS>}, \textit{i.e} a bias towards short sequences \citep{murray-chiang-2018-correcting,xuewen-etal-2021-reducing}. We found this bias to harm the recall of the tasks as the generation terminates before predicting all the entities/relations. To counteract this, we propose sentence augmentation, drawing inspiration from Pix2seq \citep{Chen2021Pix2seqAL}, who encounter a similar problem for their generative object detection model.  This approach forms an augmented training sample $s_\text{{aug}}$ by randomly concatenating sentences from the training set $\mathcal{D} = \{s_1, s_2, ..., s_N\}$:
\begin{equation}
s_\text{{aug}} = \bigoplus_{k=1}^n s_{i_k}, \quad i_k \sim \mathcal{U}(1, N)\text{, } n \sim \mathcal{U}(1, B)
\end{equation}
Here, $\mathcal{U}$ denotes the uniform distribution, and $B$ is a hyperparameter indicating the maximum number of sentences that can be concatenated. By applying this sentence augmentation technique during training, the model is exposed to diverse and longer output sequence lengths, reducing the risk of premature generation of the \texttt{<EOS>} token and thus improving recall. We perform an ablation study of the effect of sentence augmentation in our experiments section, showing that it largely improves the overall performance of our model.

\section{\label{sec:expe}Experimental Setup}
\subsection{Datasets}

\begin{table}[b]
\centering
\begin{tabular}{@{}lcccccc@{}}
\toprule
Dataset & \(|\mathcal{E}|\) & \(|\mathcal{R}|\) & \# Train & \# Dev & \# Test \\
\midrule
ACE05 & 7 & 6 & 10,051 & 2,424 & 2,050 \\
CoNLL 04 & 4 & 5 & 922 & 231 & 288 \\
SciERC & 6 & 7 & 1,861 & 275 & 551 \\
\bottomrule
\end{tabular}
\caption{The statistics of the datasets. We use ACE04, ACE05, SciERC, and CoNLL 04 for evaluating end-to-end relation extraction.}
\label{tab:dataset_statistics}
\end{table}

We evaluated our model on three benchmark English datasets for joint entity-relation extraction, namely SciERC \citep{luan2018multitask}, CoNLL 2004 \citep{carreras-marquez-2004-introduction}, and ACE 05 \citep{ace05}. The statistics of the dataset is reported on Table \ref{tab:dataset_statistics}.

\paragraph{ACE 05} is collected from a variety of domains, such as newswire, online forums and broadcast news. It provides a diverse set of entity types such as Persons (PER), Locations (LOC), Geopolitical Entities (GPE), and Organizations (ORG), along with intricate relation types that include ART (Artifact relationships), GEN-AFF (General affiliations), and PER-SOC (Personal social relationships). This dataset is particularly notable for its complexity and wide coverage of entity and relation types, making it a robust benchmark for evaluating the performance of IE models.

\paragraph{CoNLL 2004} is an annotated corpus collected
from newswires and focuses on general entities such as People, Organizations, and Locations, and relations like Work\_For and Live\_in.

\paragraph{ScIERC} is a dataset that comes with entity, coreference, and relation annotations for a collection of documents from 500 AI paper abstracts. The dataset defines scientific term types and relation types specifically designed for AI domain knowledge graph construction.

\begin{table*}[]
\renewcommand{\arraystretch}{1.}
\centering
\begin{tabular}{l|ccc|ccc|ccc}
\toprule
\multirow{2}{*}{Models}  & \multicolumn{3}{c}{SciERC} & \multicolumn{3}{c}{ACE 05} & \multicolumn{3}{c}{CoNLL 2004} \\ 
\cmidrule(lr){2-4} \cmidrule(lr){5-7} \cmidrule(lr){8-10}
& \textbf{ENT} & \textbf{REL} & \textbf{REL+} & \textbf{ENT} & \textbf{REL} & \textbf{REL+} & \textbf{ENT} & \textbf{REL} & \textbf{REL+} \\ 
\midrule
DYGIE++ \citep{wadden-etal-2019-entity}   & 67.5 & \underline{48.4} & -- & 88.6 & 63.4 & -- & -- & -- & -- \\
Tab-Seq \citep{wang-lu-2020-two}           & -- & -- & -- & 89.5 & -- & 64.3 & 90.1 & 73.8 & 73.6 \\
PURE \citep{zhong-chen-2021-frustratingly} & 66.6 & 48.2 & 35.6 & 88.7 & 66.7 & 63.9 & -- & -- & -- \\
PFN \citep{yan-etal-2021-partition}        & 66.8 & -- & \underline{38.4} & 89.0 & -- & \textbf{66.8} & -- & -- & -- \\
UniRE \citep{wang-etal-2021-unire}         & \underline{68.4} & -- & 36.9 & 89.9 & -- & 66.0 & -- & -- & -- \\ 
TablERT \citep{ma-etal-2022-joint}         & -- & -- & -- & 87.8 & 65.0 & 61.8 & \textbf{90.5} & 73.2 & 72.2 \\ 
\midrule
\multicolumn{10}{c}{\texttt{GENERATIVE}}\\
\midrule
HySPA \citep{ren-etal-2021-hyspa}          & -- & -- & -- & 88.9 & 68.2 & -- & -- & -- & -- \\
TANL \citep{paolini2021structured}         & -- & -- & -- & 89.0 & -- & 63.7 & 90.3 & -- & 70.0 \\
ASP \citep{liu-etal-2022-autoregressive} $^\dagger$  & -- & -- & -- & \textbf{91.3} & \textit{72.7} & \textit{70.5}$^\dagger$ & \underline{90.3} & -- & \textit{76.3} \\
UIE \citep{lu-etal-2022-unified}           & -- & -- & 36.5 & -- & -- & \underline{66.6} & -- & 75.0 & -- \\
LasUIE \citep{fei2022lasuie}               & -- & -- & -- & -- & -- & 66.4 & -- & 75.3 & -- \\ 
\midrule
ATG (\textit{Our model})                         & \textbf{69.7} & \textbf{51.1} & \textbf{38.6} & \underline{90.1} & \textbf{68.7} & 66.2 & \textbf{90.5} & \textbf{78.5} & \textbf{78.5} \\ 
\bottomrule
\end{tabular}%
\caption{Comparison of our proposed model with state-of-the-art methods. Results are reported in terms of Entity (ENT) F1, Relation (REL) F1, and Strict Relation (REL+) F1 scores. The best scores are shown in bold, and the second-best scores are underlined. $^\dagger$ \textit{Italic scores use undirected evaluation for relation extraction and thus are not strictly comparable to our results.}}
\label{tab:mainres}
\end{table*}

\subsection{Evaluation Metrics}
For the NER task, we adopt a span-level evaluation requiring precise entity boundaries and type predictions. To evaluate relations, we use two metrics: (1) Boundaries evaluation (\textbf{REL}) necessitates the correct prediction of entity boundaries and relation types; (2) Strict evaluation (\textbf{REL+}) additionally require accurate entity type prediction. We report the micro-averaged F1 score.

\subsection{Baselines \label{baselines}}
We succinctly and briefly describe here the baseline that we compared with our model, which we separate into two categories: Span-based/table-filling and generative IE.
\paragraph{Span-based and table-filling} \textit{DyGIE++} \citep{wadden-etal-2019-entity} is a model that uses a pretrained transformer to compute contextualized representations and employs graph propagation to update the representations of spans for prediction. \textit{Tab-Seq} \citep{wang-lu-2020-two} tackles the task of joint information extraction by treating it as a table filling problem. \textit{PURE} \citep{zhong-chen-2021-frustratingly} is a pipeline model for the information extraction task that learns distinct contextual representations for entities and relations. \textit{PFN} \citep{yan-etal-2021-partition} introduces methods that model two-way interactions between tasks by partitioning and filtering features. \textit{UniRE} \citep{wang-etal-2021-unire} proposes a joint entity and relation extraction model that eliminates the separation of label spaces for entity detection and relation classification. Their model uses a unified classifier to predict labels for each cell in a table of word pairs. In \textit{TablERT} \citep{ma-etal-2022-joint}, entities and relations are treated as tables, and the model utilizes two-dimensional CNNs to effectively capture and model local dependencies. 

\paragraph{Generative IE} \textit{HySPA} \citep{ren-etal-2021-hyspa} is a model for text-to-graph extraction that has linear space and time complexity using a Hybrid span generator. \textit{TANL} \citep{paolini2021structured} treat the joint IE task as translation from plain text to augmented natural languages by fine-tuning a T5 model \citep{Raffel2019ExploringTL}. This model has been further extended by \textit{UIE} \citep{lu-etal-2022-unified} and \textit{LasUIE} \citep{fei2022lasuie}, which both proposed better linearization and additional pretraining to enhance results. Finally, \textit{ASP} \citep{liu-etal-2022-autoregressive} handles entity and relation extraction by encoding the target structure as a series of structure-building actions, using a conditional language model to predict these actions.


\subsection{Hyperparameter Settings}
Our model, ATG, employs a transformer encoder-decoder \citep{Vaswani2017AttentionIA} architecture. We train it for a maximum of 70k steps using AdamW \citep{Loshchilov2017DecoupledWD} optimizer. We use learning rate warmup for the first 10\% of training and then decay to 0. The base learning rates are 3e-5 for the encoder, 7e-5 for the decoder, and 1e-4 for other projection layers. Unlike other generative IE models that utilize pretrained encoder-decoder architectures, often relying on large models such as T5 \citep{Raffel2019ExploringTL}, we initialize ATG's encoder with pre-trained transformer encoders, while the decoder is randomly initialized. In our preliminary experiments, we observed that initializing ATG with a pretrained encoder-decoder led to suboptimal performance. We hypothesize that this is due to our decoder's utilization of a dynamic vocabulary, whereas existing pretrained encoder-decoder models have a fixed token vocabulary, which creates a large discrepancy. We use DeBERTa \citep{He2021DeBERTaV3ID} for  Conll-04 and ACE 05 and SciBERT \citep{Beltagy2019SciBERTAP} for SciERC dataset. Across all configurations, the number of decoder layers is set to 6, though we noted that even a single layer can be enough in certain cases. The sentence augmentation hyperparameter $B$ is set to 5. 

\section{Main Results} Table \ref{tab:mainres} presents the main results of our experiments, along with comparable approaches from the literature. ATG demonstrates strong performance across all datasets. On the SciERC dataset, ATG achieves the highest scores across all metrics. It outperforms the second-best result by 0.2 in \textbf{REL+} and surpasses the best generative approach, UIE \citep{lu-etal-2022-unified}, by 2.1 points. On ACE 05, ATG provide a competitive performance, securing the second-highest scores.
The reported top-performing model, ASP \citep{liu-etal-2022-autoregressive}, operates under a relaxed, undirected relation evaluation, thereby limiting a fair comparison of results \citep{taillé2021lets}. On the CoNLL 2004 dataset, ATG exhibits its superiority by outperforming the second-best result by 2.2 in terms of \textbf{REL+}. Overall, across all three datasets, our proposed model either holds the top position or showcases strong competitive performance.

\begin{figure*}[t]
    \centering
    
    \begin{minipage}{0.32\textwidth}
        \centering
        \includegraphics[width=\linewidth]{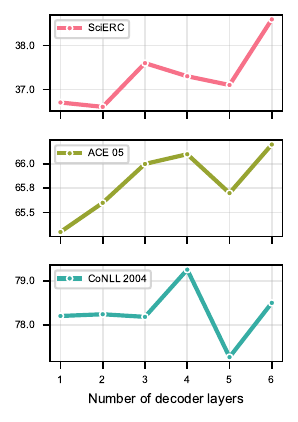}
    \end{minipage}
    \begin{minipage}{0.32\textwidth}
        \centering
        \includegraphics[width=\linewidth]{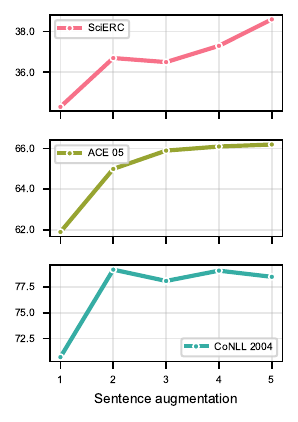}
    \end{minipage}
    \begin{minipage}{0.32\textwidth}
        \centering
        \includegraphics[width=\linewidth]{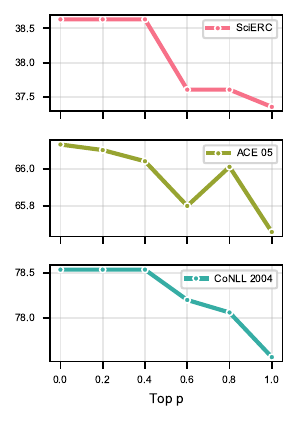}
    \end{minipage}
    
    \caption{Investigation of the effect of different choices on model performance (REL+). (\textit{Left}) Effect of the number of decoder layers, (\textit{Center}) Impact of Sentence Augmentation, (\textit{Right}) Study of different values of top-p for Nucleus Sampling.}

    \label{fig:combined}
\end{figure*}

\section{Ablation Studies}
\paragraph{Number of decoder layers} The number of decoder layers impact is illustrate on the Figure \ref{fig:combined}. It has a varying impact on performance across the datasets. In SciERC, increasing the number of decoder layers leads to a gradual improvement in the performance, reaching a peak of 38.6 at 6 layer. For ACE 05, the score shows a slight improvement from 65.3 to 66.2 as the number of decoder layers increases from 1 to 6. For the CoNLL 2004 dataset, the score fluctuates with different numbers of decoder layers, achieving already strong performance with only a single layer. Overall, the choice of the number of decoder layers can have a noticeable impact on \textbf{REL+} performance but the effect may vary across datasets.

\paragraph{Sentence augmentation} The effect of sentence augmentation size on \textbf{REL+} performance is illustrated in Figure \ref{fig:combined}. The results reveal that increasing the number of sentence augmentations always improves performance across all datasets, except for CoNLL, where achieving state-of-the-art (SOTA) results is possible with just a size of 2. However, the absence of sentence augmentation leads to a significant decrease in \textbf{REL+}, proving its importance.

\paragraph{Nucleus sampling} The impact of different top p values in nucleus sampling \citep{Holtzman2019TheCC} on the performance (REL+) is shown in Figure \ref{fig:combined}. The scores across all datasets demonstrate a relatively stable trend, with minor variations observed as the top p value changes. This can be attributed to the application of constrained decoding, which ensures that the output remains well-formed. However, the lowest values of top p, corresponding to greedy decoding, consistently deliver the best performance.

\begin{table}[!b]
\centering
\renewcommand{\arraystretch}{1.1}
\begin{tabular}{lrrr}
\toprule
         & SciERC & ACE 05 & CoNLL 2004 \\ 
\midrule
Full      & \textbf{38.6}   & \textbf{66.2}   & \textbf{78.5}       \\
- \textit{Pos}    & \underline{36.4}   & \underline{66.0}   & 78.3       \\
- \textit{Struct} & 36.1   & 65.8   & \underline{78.4}       \\
- \textit{Both}   & 35.4   & 65.4   & 78.0      \\ 
\bottomrule
\end{tabular}
\caption{Effect of positional (\textit{Pos}) and structural embedding (\textit{Struct}) on REL+. }
\label{pos_struct}
\end{table}

\begin{figure}[]
\centering
\includegraphics[width=0.8\columnwidth]{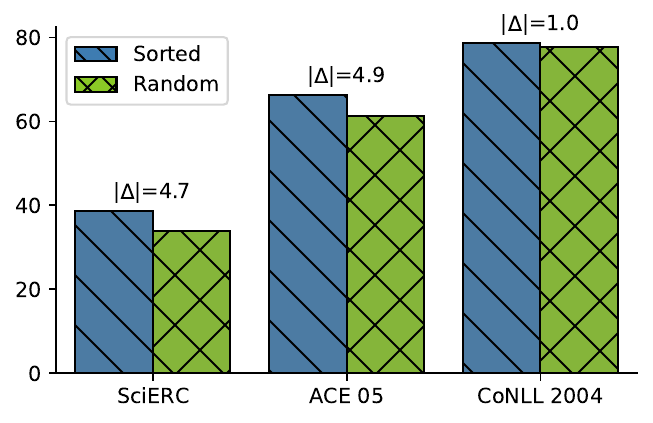}
\caption{Impact of sequence ordering on REL+.}
\label{fig:order}
\end{figure}

\paragraph{Positional and structural smbeddings} Table \ref{pos_struct} illustrates the importance of positional and structural encoding on the Relation F1 score. When employing both encoding, \textbf{ATG} achieves the best performance across all datasets (38.6 for SciERC, 66.2 for ACE 05, and 78.5 for CoNLL 2004). Excluding positional encoding causes only slight performance drops, since the span representations may contain some positional information. Omitting structural encoding leads to similar, but slightly larger drops. Finally, when both  are removed, the scores decrease the most, indicating their importance for the task.

\paragraph{Sequence ordering} Figure \ref{fig:order} compares the effects of sorted and random sequence ordering across different datasets. The results clearly show that the sorted ordering approach consistently outperforms the random one. The difference in performance is particularly significant on SciERC and ACE 05, with improvements of 4.7 and 4.9, respectively. On the CoNLL 04 dataset, although the sorted approach still leads, the difference narrows to 1. Interestingly, we initially hypothesized that random ordering would deliver better performance, given that any generation errors in a sorted order could be difficult to rectify.

\section{Interpretability Analysis}

\subsection{Attention Maps} 
\begin{figure}
    \centering
    \includegraphics[width=\columnwidth]{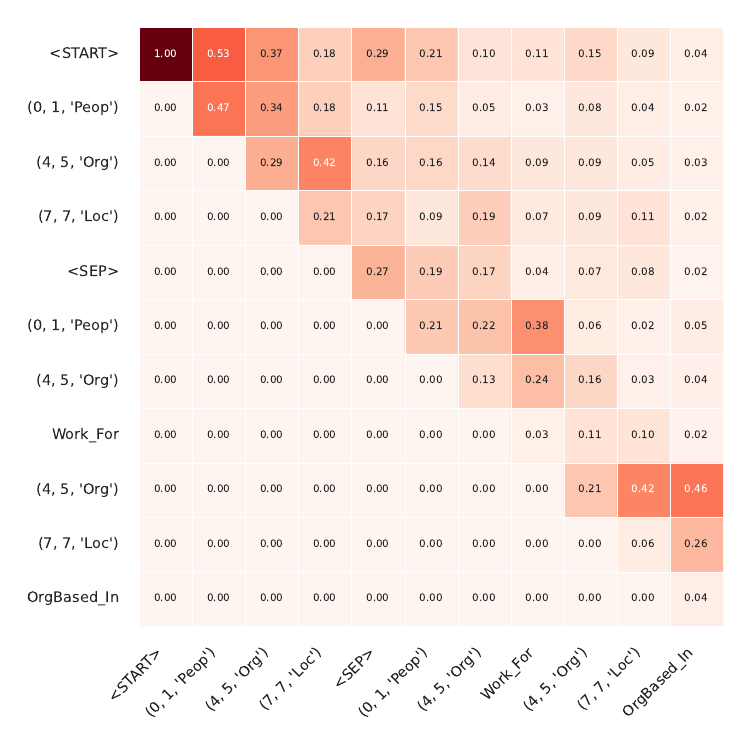}
    \caption{Decoder Self-Attention Visualization. This figure illustrates the attention patterns among elements in the generated sequence.}
    \label{fig:selfatt}
\end{figure}

\begin{figure}
    \centering
    \includegraphics[width=\columnwidth]{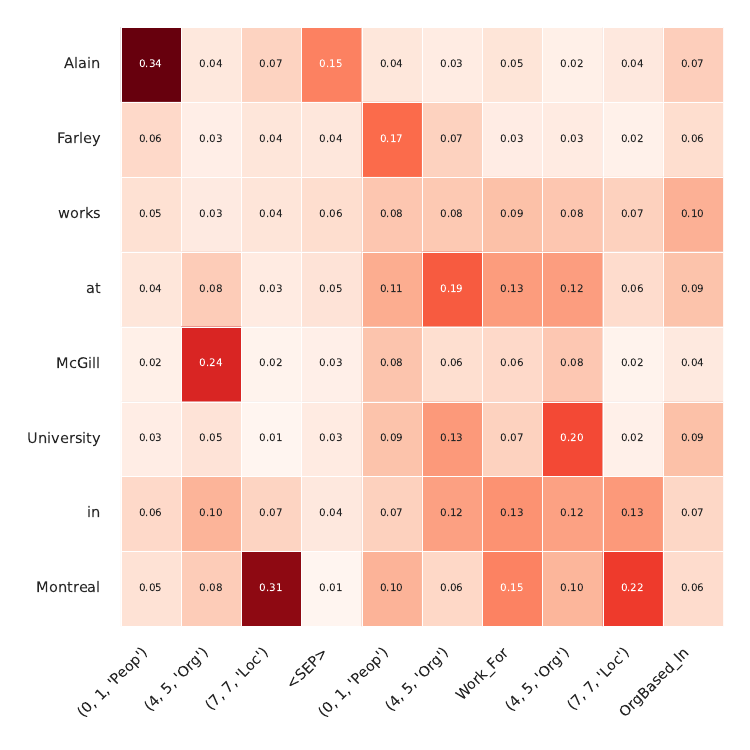}
    \caption{Decoder Cross-Attention Visualization. This map shows how each target element in the decoder interacts with and utilizes the original input text.}
    \label{fig:crossfig}
\end{figure}

Here we analyze the attention of the model during the decoding step, which allow us to explain some of the model's decision. We investigate both the self-attention (Fig. \ref{fig:selfatt}) and cross-attention (Fig. \ref{fig:crossfig}).

\paragraph{Self-attention} The self-attention map, shown in Figure \ref{fig:selfatt}, depicts the distribution of attention across preceding tokens during generation. One notable observation is the model's tendency to focus on the head and tail entities that comprise the relation when predicting relation types. For example, when predicting the \texttt{Work\_For} relation, the model allocates most of its attention weight to the tokens $\texttt{(0,1,Peop)}$  and $\texttt{(4,5,Org)}$.

\paragraph{Cross-attention} The cross-attention map in Figure \ref{fig:crossfig} indicates the specific areas in the input sequence that the decoded tokens attend to during generation. For entity labels in the output sequence such as \texttt{(0,1,Peop)}, \texttt{(4,5,Org)}, and \texttt{(7,7,Loc)}, we can observe higher attention scores for the words \texttt{Alain}, \texttt{McGill}, and \texttt{Montreal}, respectively, in the input sequence. This indicates that the model tends to focuses on the beginning of each entity span when generating these entities in the output sequence. 
Furthermore, when predicting tail entities for relations, significant attention is directed toward the prepositions 'at' and 'in' in the input sequence. This suggests that the model has learned to associate these prepositions with specific relations between entities.

\subsection{Learned Structure Embedding}
\begin{figure}[!h]
    \centering
    \includegraphics[width=.8\columnwidth]{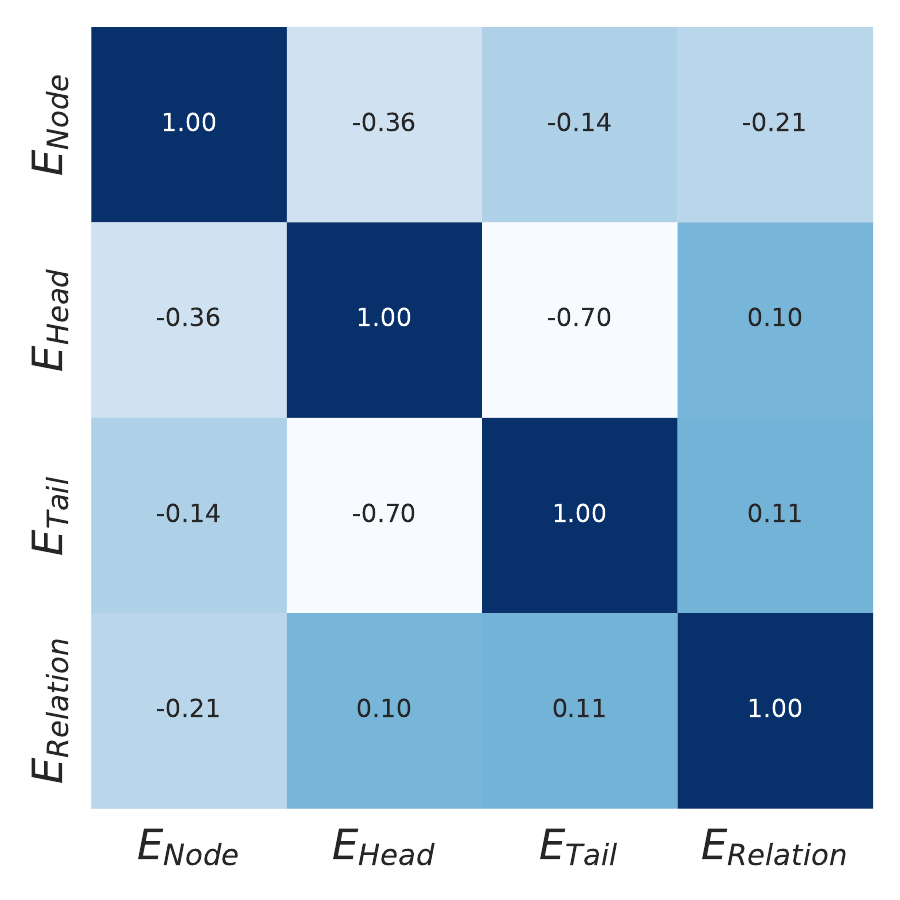}
    \caption{Structure embedding similarity. This map shows the cosine similarity between pairs of structure embedding.}
    \label{fig:struct-viz}
\end{figure}

To investigate the impact of the learned structure embedding on model performance, we analyze the similarity between the structure embeddings learned during training, depicted in Figure \ref{fig:struct-viz}. Notably, we observe a consistent pattern across datasets: the embeddings for the Head and Tail exhibit a high negative correlation. This finding may suggest that the model learns to differentiate between the Head and Tail entities, capturing their distinct characteristics.

\begin{figure}[]
    \centering
    \includegraphics[width=1\columnwidth]{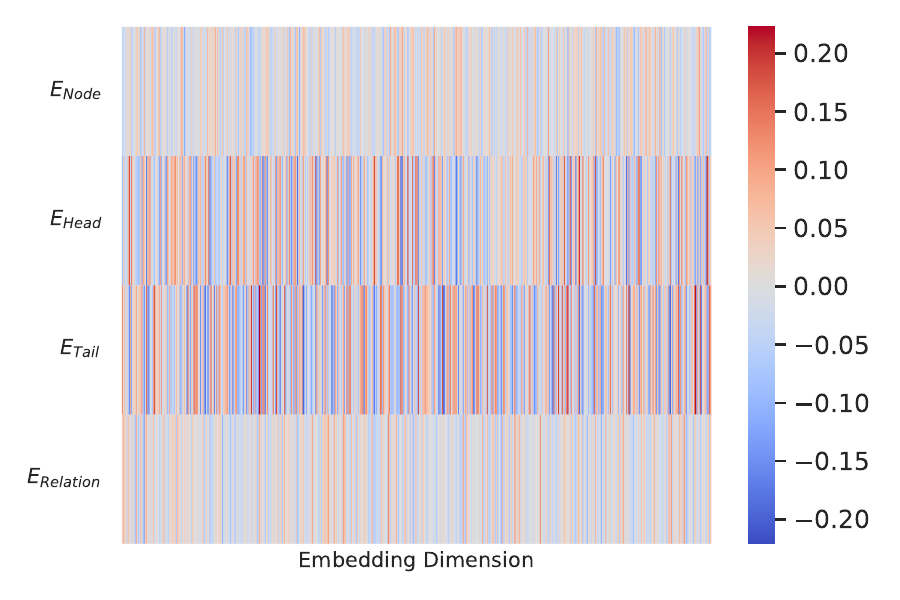}
    \caption{Structure embedding values. This shows the values taken by the learned structure embeddings.}
    \label{fig:viz-emb}
\end{figure}

However, we do not have a clear interpretation of this phenomenon. Moreover, we also report structure embedding values (over the 512 embedding dimensions) in Figure \ref{fig:viz-emb}. We observe that the structure embeddings $E_{Head}$ and $E_{Tail}$ exhibit higher values than the others, which may suggest that predicting the head and tail entities is the most challenging for the model.

\section{Related Works}
\subsection{Classification-based IE} In the field of information extraction (IE), traditional pipeline models have been used, consisting of separate stages for entity recognition and relation extraction \citep{roth-yih-2004-linear}. Entity recognition is performed to identify mentioned entities \citep{Chiu2015NamedER, Lample2016NeuralAF, zaratiana-etal-2022-gnner,zaratiana-etal-2022-named, zaratiana-etal-2023-filtered}, followed by relation extraction to determine the relationships between these entities \citep{Zelenko2002KernelMF, Bach2007ARO, Lin2016NeuralRE, Wu2017AdversarialTF}. However, this approach suffers from error propagation, where mistakes in entity recognition can negatively impact the accuracy of relation extraction \citep{10.1007/10704656_11, roth-yih-2004-linear,Nadeau2007ASO}. To address these challenges, there has been a shift towards end-to-end models that jointly optimize both entity recognition and relation extraction. This joint optimization aims to harness the interplay between the two tasks, thereby enhancing overall performance \citep{ Sun_Zhang_Mensah_Mao_Liu_2021, Zhao_Yan_Cao_Li_2021, ye-etal-2022-packed}. Noteworthy directions in this domain include table-filling methods \citep{wang-lu-2020-two, ma-etal-2022-joint}, span pair classification \citep{Eberts2019SpanbasedJE, wadden-etal-2019-entity}, set prediction \citep{Sui2020JointEA}, augmented sequence tagging mechanisms \citep{ji-etal-2020-span}, fine-grained triplet classification \citep{Shang2022OneRelJE}, and the use of unified labels for the task \citep{wang-etal-2021-unire}.

\subsection{Generative IE} Recent advancements in generative Information Extraction (IE) emphasize the use of language models (LMs) to produce entities and relations, either as text or as a sequence of actions \citep{paolini2021structured, lu-etal-2022-unified, DBLP:conf/aaai/NayakN20, liu-etal-2022-autoregressive, fei2022lasuie, wan-etal-2023-gpt}. Typically, these models employ pretrained encoder-decoder architectures, such as T5 \citep{Raffel2019ExploringTL} or BART \citep{lewis-etal-2020-bart}, to encode an input text and subsequently decode it into a structured output. Their primary advantage over non-generative methods is their ability to seamlessly integrate tasks by treating them as a unified generation process. A comprehensive review of this approach is available in \citep{ye-etal-2022-generative,xu2023large}. Beyond entity and relation extraction, generative models have also found applications in other IE tasks, including entity linking \citep{cao2021autoregressive}, event extraction \citep{li-etal-2021-document}, and document-level relation extraction \citep{Giorgi2022ASA}.

\subsection{Constrained Decoding}
In the generative IE paradigm, the model can in principle generate any sequence over the LM's vocabulary if the decoder is not constrained in some way. One might resort to \textit{controlled} to bias the model and guide the generation \citep{Li23-neurips-diffusion-lm,kumar-etal-2022-gradient,amini2023structured}. These approaches still generate a sequence over LM's vocabulary that needs to be mapped to an output graph using some kind of a \textit{parser} that analyzes the output and extracts a well-formed structure \citep{paolini2021structured}.
Another approach incorporates constraints explicitly into the decoding algorithm to restrict the LM's vocabulary to allowed tokens. For instance, \citet{cao2021autoregressive} and \citet{josifoski-etal-2022-genie} use constrained beam search to force the output to a set of allowed entities and relations from a knowledge base schema.
Another solution is to build a custom decoder that is constrained to a tailored vocabulary and decoding algorithm guaranteed to produce a linearized well-formed structure. \citet{liu-etal-2022-autoregressive} and \citet{lu-etal-2022-unified} use a specialized decoder with some sort of explicit grammar over a specific vocabulary to restrict the output to valid sequences. Our work falls in this category which can be formalized as generating words from a formal language described by a grammar over a specialized alphabet \citep{willard2023efficient,geng-etal-2023-grammar}.

\section{Conclusion}
In conclusion, our autoregressive text-to-graph framework for joint entity and relation extraction has demonstrated its effectiveness in achieving state-of-the-art or competitive results on multiple benchmark datasets. By directly generating a linearized graph representation instead of plain text, ATG successfully captures the structural characteristics, boundaries, and interactions of entities and relations. Moreover, the pointing mechanism on dynamic vocabulary provides robust grounding in the original text, which also allows our model's decoding to be fully controllable.

\section*{Acknowledgments}  This work was granted access to the HPC resources of IDRIS under the allocation 2023-AD011014472 made by GENCI.

\bibliography{aaai24}
\end{document}